\documentclass{article}

\usepackage{microtype,soul}
\usepackage{graphicx}
\usepackage{subfigure}
\usepackage{booktabs} % for professional tables
\usepackage[dvipsnames]{xcolor}
\usepackage{hyperref}

\usepackage[accepted]{icml2023}

\usepackage{amsmath}
\usepackage{amssymb}
\usepackage{mathtools}
\usepackage{amsthm}

\usepackage[capitalize,noabbrev]{cleveref}

\theoremstyle{plain}

\theoremstyle{definition}

\theoremstyle{remark}

\usepackage[textsize=tiny]{todonotes}
\usepackage{xcolor}

\icmltitlerunning{A Benchmark for Adversarial Robustness of Pruned DNN Models}%Submission and Formatting Instructions for AdvML-Frontiers 2023

\begin{document}

\twocolumn[
% \icmltitle{Benchmarking Pruned Adversarially Robust Neural Networks
%  \\
%            AdvML-Frontiers 2023}
\icmltitle{Benchmarking Adversarial Robustness of Compressed Deep Learning Models}

%Establishing a Benchmark for Adversarial Robustness of Compressed Deep Learning Models after Pruning}


\icmlsetsymbol{equal}{*}

\begin{icmlauthorlist}
\icmlauthor{Brijesh Vora}{equal,yyy}
\icmlauthor{Kartik Patwari}{equal,zzz}
\icmlauthor{Syed Mahbub Hafiz}{yyy}
\icmlauthor{Zubair Shafiq}{yyy}
\icmlauthor{Chen-Nee Chauh}{zzz}
\end{icmlauthorlist}

\icmlaffiliation{yyy}{Department of Computer Science, University of California, Davis}

\icmlaffiliation{zzz}{Department of Electrical and Computer Engineering, University of California, Davis}

\icmlcorrespondingauthor{Brijesh Vora}{bhvora@ucdavis.edu}
\icmlcorrespondingauthor{Kartik Patwari}{kpatwari@ucdavis.edu}

\icmlkeywords{Machine Learning, ICML}

\vskip 0.3in
]



\printAffiliationsAndNotice{\icmlEqualContribution} 
% otherwise use the standard text.

\begin{abstract}
% With the increase in the model sizes, parameters, and complexity the compression of these models becomes important for edge devices. Also, deep neural networks(DNNs) are vulnerable to adversarial attacks which are intelligently added perturbations to the benign examples to fool the neural network. A significant amount of work has been devoted to study the compression of the models and adversarial robustness separately but not together. This paper proposes a benchmark for compression especially filter pruning with adversarial robustness with different families of models and different white-box attacks.
% We find that at 50\% L2 pruning the pruned models are as adversarially robust as the base models. Furthermore, we show the transferability of these attacks on cross family of models and show that some pruned models are as robust if not more than base models.

The increasing size of Deep Neural Networks (DNNs) poses a pressing need for model compression, particularly when employed on resource-constrained devices. 
Concurrently, the susceptibility of DNNs to adversarial attacks presents another significant hurdle. Despite substantial research on both model compression and adversarial robustness, their joint examination remains underexplored. 
Our study bridges this gap, seeking to understand the effect of adversarial inputs crafted for base models on their pruned versions. 
To examine this relationship, we have developed a comprehensive benchmark across diverse adversarial attacks and popular DNN models. 
We uniquely focus on models not previously exposed to adversarial training and apply pruning schemes optimized for accuracy and performance.
Our findings reveal that while the benefits of pruning -- enhanced generalizability, compression, and faster inference times -- are preserved, adversarial robustness remains comparable to the base model. 
This suggests that model compression while offering its unique advantages, does not undermine adversarial robustness.
%Our investigation indicates that the adversarial robustness of models neither significantly deteriorates nor improves post-pruning.

% Previous
% work has largely focused on pruning models that
% have already undergone adversarial training and
% have demonstrated robustness. However, in cases
% where prior adversarial robustness training may
% not be feasible, it becomes critically important
% to understand the impact of adversarial attacks
% on pruned models. 

% add pipeline contributions

% We built an benchmarking pipeline/ suite 
% 

\end{abstract}

\section{Introduction}
Deep neural networks have continued to exhibit impressive performance in various machine learning applications, including computer vision, natural language processing, object detection, and so on. 
However, the deployment of these networks on resource-limited devices presents a challenge due to their substantial memory and computational requirements~\cite{chen2020deep}. 
A potential solution to this is neural network compression via pruning~\cite{zhao2019variational}, which aims to decrease size by identifying and eliminating connections that contribute less to the network's overall performance – pruning effectively reduces the number of parameters and computations required during inference. 
This compression technique optimizes the network's efficiency by focusing resources on the most critical connections, thereby enhancing its computational speed and reducing memory requirements. 
Beyond these practical constraints, another critical concern is the risk of adversarial attacks \cite{chakraborty2018adversarial}. 
These attacks craft perturbations to input data that – while appearing benign or imperceptible to humans – can mislead a machine learning model into making incorrect predictions or classifications. 
The potential implications of successful adversarial attacks are considerable, particularly in critical applications such as autonomous driving, smart health, and fraud detection ~\cite{eykholt2018robust}. 
The increasing reliance on machine learning models in mission-critical IoT and edge devices further underscores the importance of studying the relationship between model compression and adversarial robustness.

Previous work has largely focused on pruning models that have already undergone adversarial training and have demonstrated robustness \cite{ye2019adversarial,cheng2017survey,jordao2021effect}. Their primary objective is to investigate how to compress the model without nullifying the effects of adversarial training or undermining the methods that have been implemented to enhance adversarial robustness. 
However, there are cases where prior adversarial training may not be feasible -- there is often a large computation cost of robust/adversarial training \cite{wang2020improving}. Furthermore, transferable adversarial samples have been shown to overcome adversarial training \cite{tramer2017space}. Therefore, it becomes important to understand the impact of adversarial attacks under scnearios where robustness is not already guaranteed.  
Other works have also shown the effect pruning and compression can have on improving model generalization~\cite{jin2022pruning}. Our focus is on understanding the effects of pruning dense models that have not been adversarially trained or that have not undergone any adversarial robustness enhancement.

In this paper, we establish a comprehensive benchmark to evaluate adversarial robustness in pruned convolution neural networks (CNNs). Our aim is to provide a detailed understanding of the effects of existing pruning methods optimized for accuracy on models that are not already offering adversarial robustness. We consider a range of adversarial attacks (more details in Section~\ref{sec:methodology-advattacks}).

% including the Fast Gradient Sign Method (FGSM) \cite{goodfellow2014explaining}, DeepFool (DF) \cite{moosavi2016deepfool}, Projected Gradient Descent (PGD) \cite{madry2017towards}, Basic Iterative Method (BIM) \cite{kurakin2018adversarial}, Auto Projected Gradient Descent (APGD) \cite{croce2020reliable}, Carlini-Wagner (CW) \cite{carlini2017towards}, and the Universal Perturbation (UP) attack \cite{moosavi2017universal}.

We generate adversarial inputs using multiple attacks on the full and dense models, hereafter referred to as `base' models. Subsequently, we evaluate the effectiveness of these adversarial inputs on various pruned versions of the respective base models. We consider this a realistic threat model, given that large and densely trained models such as ResNets~\cite{he2016deep} are widely and publicly available in terms of both architecture and weights. These can be utilized by an attacker as surrogate models for the adversarial example crafting procedure.

Our results reveal that the pruning process has a negligible impact on the adversarial accuracy of the models. More specifically, the adversarial robustness of these models neither significantly deteriorates nor improves post-pruning while providing the benefits of pruning -- increased inference speed and better generalizability. We further extend our investigations to explore the transferability of adversarial examples across different model architectures/families. In these experiments, adversarial examples are generated from a base model architecture or family and are then fed to other base models and their pruned counterparts. These tests exhibit the same pattern as earlier, reinforcing our findings -- the adversarial impact on pruned models aligns closely with that of their base models.

\section{Background and Related Works}

\subsection{Adversarial Attacks}
Adversarial (evasion) attacks aim to craft input samples that cause misclassification by models while appearing visually similar to the original input. Adversarial attacks have continued to evolve, starting with the Fast Gradient Sign Method (FGSM)~\cite{goodfellow2014explaining} and advancing to multi-step iterative methods like Projected Gradient Descent (PGD)~\cite{madry2017towards} and optimization-based attacks like Carlini Wagner (CW) \cite{carlini2017towards}. Attacks have been rapidly growing since, as well as reflecting a dynamic cycle: as attacks grow more sophisticated, defenses adapt in response, driving swift progress in the field on both attack and defense fronts. Recent works have introduced more complex iterative methods and optimization-based attacks~\cite{dong2018boosting, croce2020reliable, xu2020generate, wang2021boosting, chen2018ead, wong2019wasserstein, ghiasi2020breaking}. Novel attacks such as adversarial patch attacks~\cite{liu2018dpatch} and adversarial examples in the physical world~\cite{dong2022isometric} have also emerged. Advancements have also been seen in enhancing attack robustness through adversarial transferability~\cite{guo2019simple, chen2020hopskipjumpattack, andriushchenko2020square} and utilization of Generative Adversarial Networks (GANs) for attack generation~\cite{xiao2018generating, mao2020gap++}.

\subsection{Neural Network Pruning}

The goal of network pruning is to eliminate redundant or unimportant connections and parameters from a neural network while maintaining or improving its performance, with techniques applied before, after, or even during training.

Post-training pruning techniques remove connections or filters based on their magnitude or contribution to the output, applied either once (single-shot) or iteratively \cite{han2015deep,liu2017learning,he2017channel,yu2018nisp}. One-shot pruning aims to remove a large portion of the network in a single step \cite{molchanov2017variational,liu2018rethinking} whereas iterative pruning involves pruning a small portion of the network at a time, then retraining the remaining part of the network \cite{tan2020dropnet, chijiwa2021pruning, han2015learning}. 

Pre-training pruning is a technique applied before training the model where the objective is to initialize a smaller network that can be trained from scratch. The lottery ticket hypothesis \cite{frankle2018lottery} introduced the concept of ``winning tickets'' in neural networks, which are subnetworks that can be trained in isolation to achieve comparable performance to the original network. Since then, various works have built upon this idea \cite{frankle2019stabilizing, evci2022gradient, frankle2020linear}

\subsection{Pruning Adversarially Robust Networks}

There has been a growing interest in pruning techniques that sustain the robustness of adversarially trained neural networks. Ye et al. \cite{ye2019adversarial} proposed a joint loss function comprising the compression rate and a robustness term, which guided the pruning of weights with the lowest $L_1$ norm. Sehwag et al. \cite{sehwag2020hydra} introduced a strategy that jointly optimizes the network's accuracy and adversarial robustness during pruning, achieved by adding a robustness-encouraging regularization term. Bai et al. \cite{bai2021improving} developed a Channel-wise Activation Suppressing (CAS) strategy to enhance a network's adversarial robustness by suppressing redundant activation based on their observation of uniform channel activation by adversarial samples. Lim et al. \cite{lim2021robustness} presented a robustness-aware filter pruning algorithm that prunes convolution layer filters based on their robustness contribution, calculated by the network's output sensitivity to each filter's removal. Lastly, Li et al. \cite{li2022can} proposed a pruning algorithm focused on neuron instability as an adversarial perturbation sensitivity metric, removing the most unstable neurons to maintain robustness.

% \cite{sehwag2019towards}
\begin{figure}[h]
% \vskip 0.2in
\begin{center}
\includegraphics[width=5.5cm]{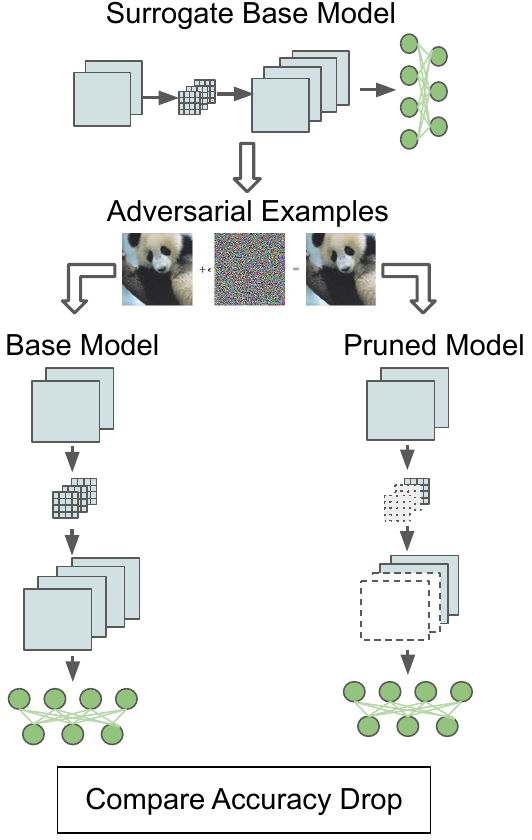}
\caption{Benchmark pipeline. Adversarial examples generated by various attacks from (attacker's) surrogate base models and evaluated on (victim's) base and pruned model.}
\label{method_fig}
\end{center}
\vspace*{-.2cm}
\end{figure}

Investigating the interplay between pruning and adversarial robustness is crucial. This exploration extends beyond adversarially trained base models, which may not always be feasible or available due to large computation demands \cite{ zhang2019theoretically, madry2017towards} and difficulty of mining samples for robust training \cite{shrivastava2016training}. Hence, we explore how adversarial robustness is affected while pruning conventionally trained models. The benchmarks we present are unique and unexplored in the literature. 

\section{Methodology}

\subsection{Threat Model}
Our benchmark pipeline, which is indicative of our threat model, is illustrated in Figure \ref{method_fig}.

\textbf{Attacker's Goal:} In adversarial attacks, the attacker can have multiple goals, being an untargeted or targeted attack. In a targeted attack, the attacker manipulates an input to make the victim model predict a specifically chosen incorrect class. Conversely, an untargeted adversarial attack aims to induce any incorrect classification without targeting a specific wrong class. Similar to the prior recent adversarial robustness benchmark (on vision transformers) by Mahmood et al. \cite{mahmood2021robustness}, we consider the untargeted attack scenario.

\textbf{Attacker's Knowledge:} We consider a white-box adversary model, which is often chosen for benchmarking adversarial attacks \cite{mahmood2021robustness, dong2020benchmarking}. In the white-box attack scenario, the adversary typically has full knowledge of the victim model's architecture and parameters. However, in our relaxed threat model, we assume the adversary lacks knowledge of the victim's trained base model parameters. Instead, the adversary can train a surrogate base model on the same dataset to generate adversarial examples. 

% \label{adversarial-attack-details}
\textbf{Attacker's Capability:} The attacker can manipulate inputs by adding perturbations (noise), which are small enough to be imperceptible by human inspection. All attacks we analyze typically confine adversarial perturbations within the bounds of the $l_p$ norm, which is standard. Details about further attack parameters, such as epsilon and iteration values, are provided under Appendix \ref{adversarial-attack-details}.

\subsection{Base Models}

For our base models, we choose well-known and readily available models, including ResNet50 (RN50) \cite{he2016deep}, DenseNet121 (DN121) \cite{huang2017densely}, VGG19 (VGG19) \cite{simonyan2014very}, and MobileNetV1 (NM) \cite{howard2017mobilenets}. While the VGGs, ResNets, and DenseNets models are large with millions of parameters, MobileNetV1 has been designed to be lightweight and less dense, providing an interesting point of comparison. Our choice of CNNs from the time and resources in our arsenal. Also, it is worth noting that the time it takes to generate adversarial examples is a few hours to days(specification mentioned in Section \ref{filter-pruning}. Therefore, we test the adversarial robustness of the pruned versions of these base models under the assumption that the adversary has access to the base model to craft inputs. We train these models on two standard datasets – CIFAR-10 and CIFAR-100 \cite{krizhevsky2009learning}.

\subsection{Adversarial Attacks}
\label{sec:methodology-advattacks}

% \label{adversarial-attack-details}
We consider a subset of popular adversarial attacks that have previously been used to benchmark adversarial robustness \cite{mahmood2021robustness,dong2020benchmarking}. Hence, our benchmark includes FGSM \cite{goodfellow2014explaining}, DeepFool (DF) \cite{moosavi2016deepfool}, PGD \cite{madry2017towards}, Basic Iterative Method (BIM) \cite{kurakin2018adversarial}, Auto Projected Gradient Descent (APGD) \cite{croce2020reliable}, and CW \cite{carlini2017towards}. We also include Universal Perturbation (UP) attack \cite{moosavi2017universal}, which is a popular attack not considered in the two prior benchmarks. We utilize the Adversarial Robustness Toolbox (ART) library to run the attacks and keep default parameters for each attack provided by ART (see Appendix \ref{adversarial-attack-details} for details). 

%%% Attack parameters
% The maximum iteration used for the CW is 10 and for DF is 5. The epsilon (Maximum perturbation that the attacker can introduce) for the FGSM is 0.2, for BIM is 1 and the eps\_step(Attacks step size at each input variation) is 0.1. For APGD epsilon is 0.2 and eps\_step is 0.1 and the maximum iteration is 5. \colorbox{orange}{Do we need to include these parameter details here??}

We generate a unique adversarial test set for each attack using base models – that is, we transform the benign CIFAR-10/CIFAR-100 test set into adversarial samples for every distinct attack. We evaluate robustness by comparing the model's accuracy on the benign test set and each adversarial test set, observing the change in performance.

\definecolor{darkgreen}{RGB}{0, 180, 0}
\begin{table}[h]
\caption{Base \& pruned model  accuracy on benign CIFAR-10 test set. The pruning target ranges from 10\% - 50\% with $L_1, L_2,$ and Geometric Median (GM) criterion. \textcolor{darkgreen}{Green} marked numbers represent the highest achieved accuracy for the specified base model across all pruning specifications, while \textcolor{red}{Red} represents the lowest. Here, max $\delta =$ max pruned accuracy $-$ base accuracy and min $\delta =$ min pruned accuracy $-$ base accuracy. \vspace*{.25cm}} 
\label{cifar10-pruning-type-without-imagenet}
\centering
% \vskip 0.15in
% \begin{center}
% \begin{small}
% \begin{sc}
\begin{tabular}{ lccccr } 
 \hline
 \toprule
   Pruning  & Pruning 	& \multicolumn{4}{c}{Benign Test Accuracy } \\
    Type & \% & MN	& DN121 &	RN50 & VGG19 \\ \midrule
   
    Base&	0\%&	79.2&	83.3&	78.1&	79.6\\	\midrule
    L2&	10\%&	\textbf{83.7}&	\textbf{86.9}&	80.6&	\textbf{82.1}\\
    L2&	20\%&	83.5&	86.2&	79&   81.6\\
    L2&	30\%&	81.8&	86.7&	\textbf{82}&	    81.3\\
    L2&	40\%&	81.1&	85.2&	74.9&	81.1\\
    L2&	50\%&	80.0&	    81.3&	70.4&	82.1 \\\midrule
    L1&	10\%&	\textbf{84.4}&   \textbf{87}&	     82.1&	\textbf{82.2}  \\
    L1&	20\%&	82.7&	85.5&	 \textbf{82.2}&	81.9\\
    L1&	30\%&	82.7&	84.6&	 78.9&	81.7\\
    L1&	40\%&	81.4&	84.3&	80.8&	81.6\\
    L1&	50\%&	80.2&	82&	    78.1&	79.7\\ \midrule

    GM&	10\%&	83.7&	86.6 &	78.6&	81.6\\
    GM&	20\%&	82.5&	86.2&	81.5&	81.5\\
    GM&	30\%&	81.4&	85.8&	83.1&	\textbf{82.5}\\
    GM&	40\%&	81.2&	84.4&	79.2&	80.3\\
    GM&	50\%&	80.3&	\textbf{80.1}&	78.6&	80.1  \\\midrule
    max $\delta$	& & 	 \textbf{5.2} &	\textbf{3.7}&	 \textbf{5}&	 \textbf{2.9} \\
    min $\delta$    & &  \textbf{0.8}&	-3.2&	-7.7& \textbf{0.1}\\
\bottomrule
\end{tabular}
% \end{sc}
% \end{small}
% \end{center}
% \vskip -0.1in
% \vspace*{-.35cm}
\end{table}

\subsection{Pruning with NNCF}

% Our study focuses on Iterative Magnitude Pruning (IMP), which is the most effective and popular pruning scheme \cite{}. During the IMP process, weights with magnitudes below a specified threshold are pruned (determined by a pruning critereon). The threshold for pruning can be determined by setting a predefined sparsity level or by selecting a specific percentage of weights with the lowest magnitudes. Following the pruning phase, the pruned network undergoes retraining to restore and optimize its performance.

% We utilize the Neural Network Compression Framework (NNCF) library \cite{kozlov2021neural} to prune base models. From NNCF, we use the filter pruning algorithm, which identifies and removes output filters in convolutional layers based on a filter importance criterion. We create pruned models using the L1, L2, and geometric median filter importance criterion. 

\label{nncf-details}
Our study concentrates on the popular and effective scheme of Iterative Magnitude Pruning (IMP)~\cite{zullich2021speeding}. In IMP, weights beneath a specified magnitude threshold, determined by a pruning criterion, are pruned. This threshold can be set by a predefined sparsity level or a specific percentage of weights with the lowest magnitudes. We use the Neural Network Compression Framework (NNCF) library \cite{kozlov2021neural} to prune base models, employing its filter pruning algorithm. This algorithm iteratively identifies and removes output filters in convolutional layers with the lowest importance, based on filter importance criteria of $L_1, L_2,$ and geometric median. Pruning targets range from 10\% to 50\% in our study (in 10\% increments), meaning up to half of the least important filters are eliminated from the network after the pruning process. Each pruning step is followed by a fine-tuning phase to optimize performance. Details about parameters for fine-tuning can be found in Appendix \ref{nncf-details}.
%---------------------------------------------------------------------------
% Cifar10 - various attacks table with L2 norm without imagenet
\begin{table}[h]
\caption{Base and pruned model on CIFAR-10. Adversarial Test Accuracies for the base model and the \textbf{L2 filter-pruned model} with 10-50\% pruning are shown. Examples are generated from the base model and fed to the base and pruned models. Bold numbers represent \emph{positive} maximum $\delta$ and the respective maximum value in the pruned model.\vspace*{.25cm}}
\label{cifar10-attacks-l2-without-imagenet}
\scriptsize
% \vskip 0.15in
% \begin{left}
% \begin{small}
% \begin{sc}
\begin{tabular}{ lccccccr } 
 \hline
 \toprule
   Attack  & Base  & \multicolumn{5}{c}{L2-Pruned}  &  max\\
     &   & 10\%	& 20\%  &30\% & 40\% & 50\% & \(\delta\) \\ \midrule
     
    MobileNet & & & & & & \\ \midrule  
    CW&	67.5&	65.8&	64.9&	65.2&	63.6&	65.2&	-1.7 \\
    DF&	40.7&	\textbf{43}&	\textbf{43}&	42.3&	41.8&	41.8&	\textbf{2.3}\\
    FGSM&	12&	12.4&	12.3&	11.3&	\textbf{12.8}&	11.1&	\textbf{0.8}\\
    BiM&	5.1	&4.1&	3.5&	3.8	&3.9	&\textbf{5.6}&	\textbf{0.5}\\
    PGD&	7.4&	4.1	&4.6&	4.2	&4.7&	6.7	&-0.7\\
    APGD&	8.5	&3.7&	4.1	&4.4&	5.8	&8&	-0.5\\
    UP&	58.8&	\textbf{64.9}&	63	&60.1	&60.3&	59	&\textbf{6.1} \\\midrule
    
    DenseNet121 & & & & & & \\ \midrule
    CW&	72.1&	70&	68.1&	69.9&	69.1&	66.9&	-2.1 \\
    DF&	38.7&	38.1&	38.2&	\textbf{38.8}	&37.8&	35.6&	\textbf{0.1}\\
    FGSM&	11.3&	11.6&	\textbf{12}&	10.9&	10.6&	10.3&	\textbf{0.7}\\
    BiM	&8&	6.5&	6.1&	6.7&	6.5&	6.7&	-1.3\\
    PGD	&7.1	&6.7&	6.8&	6.8&	6.7	&7.1&	0\\
    APGD&	4.5&	3.9&	4&	3.8&	3.6&	3.8&	-0.5 \\
    UP	&10.2&	10&	10.6& \textbf{11.7} &	10.3&	10&	\textbf{1.5} \\\midrule

    ResNet50 & & & & & & \\ \midrule
    CW	&68.3&	71.4&	69.4&	\textbf{72.7}&	65&	60.4&	\textbf{4.4} \\
    DF&	37.1&	39.7&	39.7&	\textbf{40.4}&	35.9&	33.8&	\textbf{3.3}\\
    FGSM&	15.4&	11.9&	13.6	&13.6&	13.2&	13	&-1.8\\
    BiM&	7&	7.2&	\textbf{7.6}&	7.4&	7.4&	6.9&	\textbf{0.6}\\
    PGD	&7.8&	7.5&	7.5	&7.5&	7.1	& \textbf{8.3}&	\textbf{0.5}\\
    APGD&	4.9&	3.9&	4.2&	4.1&	4.4&	\textbf{6.5}&	\textbf{1.6} \\
    UP&	15.6&	14.1&	11.8	&14.9&	11.4&	11&	-0.7 \\\midrule
      
    VGG19 & & & & & & \\ \midrule
    CW& 67.2&	\textbf{70.2}&	69.6&	70&	68.7&	70.1&	\textbf{3}\\
    DF&	38.9 &	\textbf{39}&	38.8&	\textbf{39}&	37.1&	38.1&	\textbf{0.1}\\
    FGSM&	17.8 &	16.9&	17.6&	17.4&	\textbf{18.1}&	16.9& \textbf{0.3}\\
    BiM	&9.2&	9.2	&9.2	&9.2&	9.1&	\textbf{9.3}&	\textbf{0.1}\\
    PGD	&7.7&	7.5&	7.8&	7.3&	\textbf{8.3}	&7.6&	\textbf{0.6}\\
    APGD&	4.4&	3.8&	3.8&	4&	4&	4.1&	-0.3 \\
    UP&	49.4	&48.7&	48.9&	\textbf{50.6}&	47.3&	48.8&	\textbf{1.2}\\ 
\bottomrule
\end{tabular}
% \end{sc}
% \end{small}
% \end{left}
% \vskip -0.1in
% \vspace*{-.3cm}
\end{table}

% Cifar10 -Inference table

\begin{table}[h]
\caption{Inference time (ms) of base and $L_2$-pruned models on CIFAR-10. Inference time is calculated averaged over 100 runs with batch size = 64.\vspace*{.25cm}}
\label{cifar10-inference-time}
\scriptsize
% \vskip 0.15in
% \begin{center}
% \begin{small}
% \begin{sc}
\begin{tabular}{lccccr} 
 \hline
 \toprule
 Model &   \multicolumn{4}{c}{Inference Time (ms)} \\
      & Base & 10\% & 30\%  & 50\%   \\ 
 \midrule  
VGG19 & 113.80 $\pm$ 7.81 &	104.95 $\pm$ 8.81&	101.13 $\pm$ 6.12&	103.65 $\pm$ 5.56  \\
RN50 	& 95.07 $\pm$ 4.82	&88.35 $\pm$ 4.29&	89.40 $\pm$ 5.71&		81.41 $\pm$ 3.82 \\

MN  &	 23.48 $\pm$ 2.46 &	 20.86 $\pm$ 2.08&  	20.39 $\pm$ 1.41& 		21.46 $\pm$ 1.97 \\
DN121	&  71.13 $\pm$ 5.27&	65.10 $\pm$ 3.80&	 61.09 $\pm$ 4.17&	60.55 $\pm$ 4.05 \\
\bottomrule
\end{tabular}
% \end{sc}
% \end{small}
% \end{center}
% \vskip -0.1in
% \vspace*{-.2cm}
\end{table}
%---------------------------------------------------------------------------

%---------------------------------------------------------------------------

\section{Results}

\subsection{Filter Pruning}
\label{filter-pruning}
Table \ref{cifar10-pruning-type-without-imagenet} shows the benign test accuracies with the base model and pruned models. We observe that the benign test accuracy for all the model families initially increases (with pruning 10\%-40\%), offering better generalizability and model performance while reducing the number of parameters. We notice this improvement is in part due to the fine-tuned training after every pruning step. At about 50\% pruning rate, we see that accuracy is generally similar to the original base model. Notably, there is one outlier in this scenario, ResNet50 – with L2 pruning up to 50\%, which has a 7.7\% drop-off. \textbf{Overall, we find that NNCF's iterative magnitude filter pruning results in compressed models with fewer parameters and better generalizability, and fine-tuning helps achieve close to the original base model performance, if not better.} Benchmarks on CIFAR-100 – presented in Appendix \ref{cifar100-results} – demonstrate similar trends. 

Table \ref{cifar10-inference-time} shows the inference time of base and various L2-pruned models. The inference results were generated on a CPU (AMD EPYC 7302 16-Core Processor $@$ 1.49GHz, 256GB of RAM). We aimed to mimic the constraints of real-world deployment scenarios where high-end server-class GPUs may not be available and their inference time performance does not truly reflect the operational conditions of constrained devices. Both the base and pruned models, initially in the .h5 format, were converted to the OpenVINO format (NNCF's preferred format) for the purpose of inferencing. We ran inference 100 times using a batch size of 64 and reported both the mean and standard deviation of the results. As expected, we find that as the models are pruned from 0\% to 50\%, the inference time decreases. 

Table \ref{cifar10-size} shows various sizes of the models after pruning. The pruned model sizes are the same for 10\% to 50\% for NNCF filter pruning (when measuring .h5 file size). From both these tables, we conclude that the size of the models reduces by around 66\% and gets a boost in the inference of about 8-10\%. \textbf{Thus, we see that filter pruning also helps reduce the model size and inference time.}

% Cifar10 -size table

%---------------------------------------------------------------------------

\begin{table}[h]
\caption{Base and pruned model sizes (MB) on CIFAR-10. \vspace*{.25cm}}
\label{cifar10-size}
\centering
% \scriptsize
\begin{tabular}{ lcccr } 
 \hline
 \toprule
 Model &  \multicolumn{4}{c}{Size (MB)} \\
Type &  MN & DN121& RN50 & VGG19   \\ 
 \midrule 
Base & 38& 82&  271& 230 \\ 
 Pruned & 13.2 & 29.4& 95 & 77\\
\bottomrule
\end{tabular}
% \vspace{-0.3in}
\end{table}

\subsection{Filter Pruning \& Adversarial Robustness}
\label{ss-adv_robustness}
Our primary goal is to understand the effect of iterative filter pruning on adversarial robustness. Table \ref{cifar10-attacks-l2-without-imagenet} presents the accuracy achieved on the CIFAR-10 adversarial test sets crafted on the base models, comparing it to the performance when the same test set is fed to their respective L2 pruned models.  We use the weights of the base models to fine-tune the pruned model after removing 10\% of the filters. CIFAR-100 results are shown in Appendix \ref{cifar100-results}.
%% change this refernce to cifar100 results subsection not table

%While adversarial robustness remains unchanged, we benefit from the advantages of compression from pruning. 

%---------------------------------------------------------------------------
% Cifar10 - various attacks table with GM norm - imagenet =false
\begin{table}[H]
\caption{Base and Pruned model on CIFAR-10. Adversarial Test Accuracy and \textbf{ GM filter pruned model} with 10- 50\%  pruning and their Adversarial Test Accuracy are shown. Examples are generated from the base model and fed to the base and pruned models.\vspace*{.25cm}}
\label{cifar10-attacks-gm-without-imagenet}
% \scriptsize
% \vskip 0.15in
\begin{center}
\begin{tiny}
% \begin{sc}
\begin{tabular}{ lccccccr } 
 \hline
 \toprule
 Attack  & Base  & \multicolumn{5}{c}{GM Pruned}  &  max\\
     &   & 10\%	& 20\%  &30\% & 40\% & 50\% & \(\delta\) \\ \midrule
     MobileNet & & & & & & & \\ \midrule
    CW	&67.5&	64.6	&66.5	&64.7&	64.3&	67.5&	0 \\
    DF&	40.7&	\textbf{44}	&42.4&	40.1&	42.2	&41.8&	\textbf{3.3}\\
    FGSM&	12	&\textbf{21.8}&	12.5&	13.6&	11.9&	12.1&	\textbf{9.8}\\
    BiM	&5.1&	\textbf{8.4}&	3.5	&3.8&	3.8&	4.4&	\textbf{3.3}\\
    PGD	&7.4	&\textbf{18.3}&	4.6	&4.6	&5	&5.8	&\textbf{10.9}\\
    APGD&	8.5	&\textbf{19.3}&	4.9	&5	&5.1&	5.6	&\textbf{10.8}\\
    UP&	58.8&	\textbf{66.9}	&63 &60.4&	57.8&	57&	\textbf{8.1}\\ \midrule

 DenseNet121 & & & & & & & \\ \midrule
    CW	& 72.1&	58.5&	69.8&	69.5&	68.3&	65.1	&-2.3\\
    DF	&38.7	&35.7&	37.9&	36.6&	35.7&	36	&-0.8\\
    FGSM&	11.3&	\textbf{24.6}	&11.6&	11.8&	11.3&	9.5&	\textbf{13.3}\\
    BiM	&8	&\textbf{8.5}&	6.1&	8&	6.4&	6.7&	\textbf{0.5}\\
    PGD	&7.1&	\textbf{18.4}&	6.8	&6.7&	6.7	&6.6	&\textbf{11.3}\\
    APGD&	4.5&	\textbf{18.6}&	3.9	&3.9&	3.7	&3.7	&\textbf{14.1}\\
    UP	&10.2&	\textbf{16.3}&	10.7&	10.4&	9.7&	10.1&	\textbf{6.1}\\ \midrule

 ResNet50 & & & & & & & \\ \midrule
    CW	&68.3&	67.4&	71.5&	\textbf{73.6}&	68.9&	67.7	&\textbf{5.3}\\
    DF&	37.1&	34.8&	37.5&	\textbf{40.6}	&38.5&	36.7&	\textbf{3.5}\\
    FGSM&	15.4&	13.7	&11.5&	14.3&	12.9&	11.4&	-1.1\\
    BiM&	7&	7.2&	\textbf{7.3}	&7.3	&7.3&	7.2	&\textbf{0.3}\\
    PGD&	7.8	&7.2&	7.4&	7.5	&7.3&	7.5	&-0.3\\
    APGD&	4.9	&4	&4&	4.4&	4.4&	4.4&	-0.5\\
    UP	&15.6&	11.7&	11.5&	10.8&	11.9&	11.3&	-3.7\\ \midrule

    VGG19 & & & & & & & \\ \midrule
    CW	&67.2	&\textbf{70.9}&	69.6&	70.4&	69.4	&66.7&	\textbf{3.7}\\
    DF	&38.9&	38.7&	\textbf{39}	&38.9	&36.3	&37	&\textbf{0.1}\\
    FGSM	&17.8&	15.6&	18.3&	\textbf{18.4}&	17.2&	17&	\textbf{0.6}\\
    BiM	&9.2&	9.2	&\textbf{9.3}&	9.3&	9.2&	9.2	&\textbf{0.1}\\
    PGD	&7.7&	7.7&	7.8&	7.5	&8.4	&\textbf{8.6}&	\textbf{0.9}\\
    APGD	&4.4&	3.9	&3.7&	3.8&	3.8&	4.2	&-0.2\\
    UP	&49.4&	49.7&	51.3	&\textbf{50.1}&	45.3&	48.3	&\textbf{1.9}\\
   
\bottomrule
\end{tabular}
% \end{sc}
\end{tiny}
\end{center}
% \vskip -0.1in
\end{table}

Our findings indicate a minimal impact on adversarial accuracy stemming from the pruning process, thus suggesting a relative consistency in the adversarial robustness of the models. Specifically, our experimental results highlight that the process of pruning neither significantly degrades nor enhances the robustness of the models when exposed to adversarial attacks. Primarily the maximum change in accuracy (max $\delta$) is between $\pm 1\%$, with some attacks being slightly higher. The most significant change occurs using UP attack on MobileNet, where the 10\% pruned MobileNet's accuracy on UP adversarial test set increases from its base model by about 6\%. Our findings underline an intriguing invariance in adversarial robustness when examples are generated from a base model and fed into its pruned models with no prior adversarial robustness measures taken. The results for GM pruned models are shown in table \ref{cifar10-attacks-gm-without-imagenet} and L1 in Appendix \ref{cifar10-results}. While the same trend applies, we notice for GM pruned models at 10\% we see a slight boost in accuracy (3\%-10\% increase) for MobileNet and DenseNet121. However, this is not a consistent trend across other models, and as the pruning target increases up to 50\%. The NNCF-based pruning models initially removes the 10\% unimportant filters and then for a target pruning of 20-50\% it iteratively fine tunes and removes the filters and weights, which leads to the PGD and APGD attack to show significant accuracy change. Furthermore, the accuracy of the CW is \( >=60\%\) because of the adversarial examples are generated using the surrogate base model using a similar training scheme. \textbf{Overall, our results show that pruning base models result in compressed models that run faster while maintaining comparable performance and adversarial robustness.} 
%---------------------------------------------------------------------------
% Cifar10 - transferabilty results
\begin{table}[h]
\caption{Transferability results on CIFAR-10. The adversarial test sets are generated from (surrogate) ResNet50 and fed to the victim models with different architectures. Accuracies shown are for victim models. Bold numbers represent \emph{positive} maximum $\delta$ and the respective maximum value in the pruned model.\vspace*{.25cm}}
\label{cifar10-transferability-main}
\scriptsize

% \vskip 0.15in
% \begin{center}
% \begin{small}
% \begin{sc}
\begin{tabular}{ lccccccr } 
 \hline
 \toprule 
 Pruning \% & \multicolumn{7}{c}{Surrogate Model: ResNet50} \\
   L2 & APGD&	BiM	&CW&	DF&	FGSM&	PGD	&UP \\\midrule

   MobileNet & & & & & & \\ \midrule
    0\%&13.5&	9.3&	78.5&	41&	12.6&	12.5&	11.7 \\
   10\%& 9.1&	9.1&	\textbf{83}&	\textbf{45}&	\textbf{13}&	9&	11.3 \\
   20\% &9.2&	9.2&	82.3&	44	&12.5&	9.6&	11.2\\
   30\% &8.7&	9&	81.6&	43&	12.8&	8.5&	11.2\\
   40\% &11.1&	9.7&	79.2&	41.4&	12.1&	10.3&	\textbf{12}\\
   50\% &10.8&	\textbf{9.9}&	78.1&	41&	10.9&	9.8	&11.4\\
    max $\delta$ &-2.4&	\textbf{0.6}&   \textbf{4.5}	&\textbf{4}&	\textbf{0.4}	&-2.2&	\textbf{0.3} \\ \midrule

    DenseNet121 & & & & & & \\ \midrule			
   0\% &10.4&	11.7&	83&	43.3&	14.3&	11.5&	11.6\\
   10\% &6.5&	8.7&	\textbf{85.7}&	\textbf{45.3}&	12.4&	8.2&	10\\
   20\% &6.8&	7.4&	84.6&	44.3&	12.4&	8.1&	10.1\\
   30\% &6.2&	8.2&	85.3&	44.4&	12.2&	8.6&	9.9\\
   40\% &5.8&	7.4&	84.8&	43.9&	12.4&	8.6&	11\\
   50\% &6.1&	7.2&	80.5&	39.6&	12.1&	9.7&	10\\
   max $\delta$ &  -3.6&	-3&	\textbf{2.7}&	\textbf{2}&	-1.9&  -1.8	&-0.6 \\ \midrule
    VGG19	& & & & & & \\ \midrule					
    0\% &7.9&	8.5&	78.5&	41.3&	17.2&	8.6	&12.4\\
    10\%&9.1&	8.5&	\textbf{81.5}&	43.5&	15.4&	8.2&	10.9\\
    20\% &\textbf{9.5}&	7.6&	80.5&	43.7&	15.5&	8.5&	12\\
    30\% &7.3&	7.6&	80.2&	41.1&	15.4&	8.2&	12.6\\
    40\% &8.6&	8.1&	80.5&	43.5&	15.6&	8.5	&\textbf{14.5}\\
    50\% &7.1&	8.2&	81.1&	\textbf{44.6}&	14.4&	8&	10.1\\
    max $\delta$ &   \textbf{1.6}&	{0}&	\textbf{3}&	\textbf{3.3}&	-1.6&	-0.1&	\textbf{2.1} \\
   
   \bottomrule
\end{tabular}

\end{table}

% \vspace*{-.3in}
\subsection{Adversarial Transferability}
\label{adversarial-transferability}

This study explores the effect of feeding adversarial examples generated from one model architecture family into different model families. This departs from the previous sections, where adversarial examples were tested within the same model family. Here, we consider adversarial test sets or examples created from the surrogate base model -- ResNet50, and these are now cross fed into all other base models and their respective pruned models. The focus of this investigation is to understand the phenomena and implications of adversarial transferability ~\cite{guo2019simple, chen2020hopskipjumpattack, andriushchenko2020square} across different pruned architectures. Table \ref{cifar10-transferability-main} demonstrates our findings when ResNet50 is used as the surrogate base model for adversarial example generation. Results for the remaining models as surrogate models demonstrate similar trends (see Appendix \ref{cifar10-transferabilty-results}). We observe that the pruned models do not show any significant variations in their adversarial transferability compared to their base models. This corroborates our primary findings from Section \ref{ss-adv_robustness}. \textbf{Even in a cross-model testing environment, the pruned models exhibit adversarial robustness comparable to their original base models.} 

\section{Conclusion and Future Work}
In conclusion, this paper presents a rigorous evaluation of the impact of filter pruning on the adversarial robustness of neural network models. We demonstrate, using the CIFAR-10 and CIFAR-100 datasets, that despite compressing the model by up to 50\% through filter pruning, the adversarial robustness remains relatively unaffected as compared to the base models. Our results offer promising implications, indicating that while practitioners can enjoy the benefits of filter pruning – such as accelerated inference time, curtailment in over-parameterization, and enhanced generalization capabilities – they do not have to compromise on adversarial resilience. Finally, our study supports the application of filter pruning, showcasing no detrimental effects on adversarial robustness with respect to the original base model. 

One crucial area for future investigation involves the exploration of alternative compression techniques that go beyond the scope of NNCF-based pruning. It is essential to expand the repertoire of compression methods to find innovative approaches that can further enhance the efficiency of neural network models. By venturing into unexplored territories, researchers can discover novel ways to compress models effectively, reducing their size and computational requirements while maintaining high performance, including comparable robustness against evasion attacks. %Therefore, future efforts should focus on investigating and developing new compression techniques, allowing for a broader range of options and advancements in the field of neural network compression. 
Furthermore, we encourage future research to continue exploring the intersection of model compression and adversarial robustness, contributing further to the creation of efficient, secure, and robust models ready for real-world deployment. 

\section*{Acknowledgements}
We thank the anonymous reviewers for their helpful comments. This project is based upon work supported in part by the UC Noyce Institute: Center for Cybersecurity and Cyberintegrity (C-CUBE).
% \textbf{Do not} include acknowledgements in the initial version of
% the paper submitted for blind review.

% If a paper is accepted, the final camera-ready version can (and
% probably should) include acknowledgements. In this case, please
% place such acknowledgements in an unnumbered section at the
% end of the paper. Typically, this will include thanks to reviewers
% who gave useful comments, to colleagues who contributed to the ideas,
% and to funding agencies and corporate sponsors that provided financial

\nocite{langley00}

\bibliography{main}
\bibliographystyle{icml2023}

%%%%%%%%%%%%%%%%%%%%%%%%%%%%%%%%%%%%%%%%%%%%%%%%%%%%%%%%%%%%%%%%%%%%%%%%%%%%%%%
%%%%%%%%%%%%%%%%%%%%%%%%%%%%%%%%%%%%%%%%%%%%%%%%%%%%%%%%%%%%%%%%%%%%%%%%%%%%%%%
% APPENDIX
%%%%%%%%%%%%%%%%%%%%%%%%%%%%%%%%%%%%%%%%%%%%%%%%%%%%%%%%%%%%%%%%%%%%%%%%%%%%%%%
%%%%%%%%%%%%%%%%%%%%%%%%%%%%%%%%%%%%%%%%%%%%%%%%%%%%%%%%%%%%%%%%%%%%%%%%%%%%%%%
\newpage
\appendix
\onecolumn
\section{Appendix}
\label{appendix}

% You can have as much text here as you want. The main body must be at most $6$ pages long.
% For the final version, one more page can be added.
% If you want, you can use an appendix like this one, even using the one-column format.

\subsection{Adversarial Attack Details}
\label{adversarial-attack-details}
% The maximum iteration used for the CW is 10, and for DF is 5. The epsilon (Maximum perturbation that the attacker can introduce) for the FGSM is 0.2, for BIM is 1, and the eps\_step(Attacks step size at each input variation) is 0.1. For APGD, epsilon is 0.2 and eps\_step is 0.1, and the maximum iteration is 5. 
\vspace{-0.3in}
\begin{table}[h]
\caption{Attack details }
\label{cifar10-attacks-details}
% \scriptsize

\begin{center}
\begin{tiny}
% \begin{sc}
\begin{tabular}{ lccr } 
 \hline
 \toprule
    Attack & Epsilon & Epsilon Step & Max iteration \\ \midrule
    CW & - & 0.01 & 10   \\
    DF & 1e-6 & - & 5   \\
    FGSM & 0.2 & & 100   \\
    BiM & 1 & 0.1 & 100 \\
    PGD & 0.3 & 0.1 & 100 \\
    APGD & 0.3 & 0.1 & 100 \\
    UP & 10 & - & 1 \\
 \bottomrule
\end{tabular}
% \end{sc}
\end{tiny}
\end{center}
% \vskip -0.1in
\end{table}
% We generate an adversarial test set per attack using base models, i.e., we take the (benign) CIFAR-10/CIFAR-100 test set and use all test set images to create adversarial samples with each attack. Hence, the adversarial test set for CIFAR-10 would be 10,000 images, with 1000 images from each class. We then evaluate robustness by measuring the accuracy drop of the model on the benign and adversarial test sets.

\subsection{Neural Network Pruning Details}
\label{nncf-details}
% The ML Pipeline consists of automation scripts that ease the process of training the models, generating adversarial samples from different attacks via the Adversarial Robust Toolbox (ART) \cite{art2018} library, getting the pruned model from Neural Network Compression Framework (NNCF) \cite{kozlov2020neural} and calculating the base and pruned model accuracy on benign and adversarial samples. 

% We use NNCF with TensorFlow for compressing our models using $L_1$ and $L_2$ norm and with Geometric Median with [10\%, 20\%, 30\%, 40\%, 50\%] pruning capacity. All the configs are written in a JSON file. The pre-trained model is used with an input of size $32\times32\times3$. 

The batch size is 128 and 10 epochs for (base) training. The models have been trained on CIFAR-10/CIFAR-100 trainset. We choose the optimizer as SGD with a learning rate of 0.1, gamma as 0.1 and steps as [10, 20, 30]. The  momentum of 0.9 and nesterov is set to true. The schedule\_step is multistep. For compression we use algorithm as filter\_pruning, and schedule as exponential. The pruning\_init is set to 0.1 for all the  $L_1$ and $L_2$ and GM but the pruning\_target varies from 0.1 to 0.5 [10 - 50\%]. The filter\_importance is [$L_1$ and $L_2$ and GM].

% For the compression framework, we use filter pruning (of NNCF) as it determines which filters are important and which to remove. Initially, we set the initial pruning level target to 10\% so that it removes 10\% of the filters. The pruning target (pruning level target at the end of schedule) to [10-50\%].  Then it fine-tunes the model to the pruning target over 15 epochs. 

% We have taken 4 models from different families ResNet50 \cite{he2015deep}, DenseNet121 \cite{huang2017densely}, MobileNet\cite{howard2017mobilenets}, VGG19\cite{simonyan2014very} and with 7 white box attacks namely Fast Gradient Sign Method (FGSM)\cite{goodfellow2014explaining}, DeepFool (DF)\cite{moosavi2015deepfool}, Projected Gradient Method (PGD)\cite{madry2017towards}, Carlini-Wagner (CW)\cite{carlini2016towards}, Basic Iterative Method(BiM)\cite{kurakin2016adversarial}, Auto Projected Gradient Descent (APGD)\cite{croce2020reliable}, and Universal Perturbation (UP) \cite{moosavi2017universal}.

%---------------------------------------------------------------------------

\subsection{CIFAR-10 Results}
\label{cifar10-results}

% \subsubsection{Pruned models finetuned using base model weights}
% \label{cifar10-results-without-imagenet}

%---------------------------------------------------------------------------
% Cifar10 - various attacks table with L1 norm - imagenet =false
\begin{table}[H]
\caption{Base and Pruned model on CIFAR-10. Adversarial Test Accuracy and \textbf{ L1 filter pruned model} with 10- 50\%  pruning and their Adversarial Test Accuracy are shown. Examples are generated from the base model and fed to the base and pruned models. }
\label{cifar10-attacks-l1-without-imagenet}
% \scriptsize
% \vspace{-2.5in}
\begin{center}
\begin{tiny}
% \begin{sc}
\begin{tabular}{ lccccccr } 
 \hline
 \toprule
  Attack  & Base  & \multicolumn{5}{c}{L1 Pruned}  &  max\\
      &   & 10\%	& 20\%  &30\% & 40\% & 50\% & \(\delta\) \\ \midrule

        MobileNet & & & & & & & \\ \midrule
   	CW&	67.5&	65.6&	65.8&	66.2&	66.2&	65.3&	-1.3\\
	DF&	40.7&	42.7&	\textbf{43.3}&	41.2&	39.4&	40.1&	\textbf{2.6}\\
	FGSM&	12&	12.3&	12.5&	\textbf{13.7}&	13.1&	12.8&	\textbf{1.7}\\
	BiM&	5.1&	4.6&	4.1&	3.8&	3.1&	3.1&	-0.5\\
	PGD&	7.4&	4.8	&4.5&	4.7&	4.6&	4.4&	-2.6\\
	APGD&	8.5&	3.8&	4&	4.4&	4.6&	4.6&	-3.9\\
	UP&	58.8&	65&	60.4&	\textbf{61.2}&	56.7&	56.6&	\textbf{6.2}\\ \midrule
  DenseNet121 & & & & & & & \\ \midrule
	CW	&72.1&	70.3&	69.1&	68.1&	67.9&	69.7&	-1.8\\
	DF&	38.7&	38&	36.2&	33.3&	34.9&	35.7&	-0.7\\
	FGSM&	11.3&	11.9&	12	&\textbf{12.7}	&11.4&	9.8	&\textbf{1.4}\\
	BiM&	8&	7.7	&\textbf{8.4}	&6.2&	6.4&	7.2	&\textbf{0.4}\\
	PGD	&7.1	&6.7	&7.1&	6.8	&6.8&	\textbf{8}	&\textbf{0.9}\\
	APGD&	4.5	&4	&4&	3.9	&3.7	&3.7&	-0.5 \\
	UP&	10.2	&10	&9.8	&10.2&	10	&10	&0 \\ \midrule
 ResNet50 & & & & & & & \\ \midrule
  	CW&	68.3&	73.1&	\textbf{72.4}&	69.2&	71.2&	68.1&	\textbf{4.8}\\
	DF&	37.1&	39.8&	\textbf{40.2}&	39.3&	38.7&	37.8&	\textbf{3.1}\\
	FGSM&	15.4&	12.4&	12.2&	13.1	&13.1&	11.9&	-2.3\\
	BiM&	7&	7.3&	7.6&	7.4&	\textbf{7.7}&	7.3&	\textbf{0.7}\\
	PGD	&7.8	&7.6	&7.5&	7.5	&7.2&	7.5	&-0.2\\
	APGD&	4.9&	4.1&	4	&3.8	&3.7&	4.7&	-0.2\\ 
	UP&	15.6 & 11.7&	15.2&	11.8	&11.9	&\textbf{17.3}&	\textbf{1.7} \\\midrule

    VGG19 & & & & & & & \\ \midrule
     CW&	67.2&	\textbf{70.4}&	70.4&	69.2&	69.6&	67.6&	\textbf{3.2} \\
        DF	& 38.9&	38.8&	\textbf{39.2}&	38.9&	37.2&	38.8&	\textbf{0.3} \\
        FGSM&	17.8&	16.5&	18&	17.8&	\textbf{18.8}&	15.3&	\textbf{1}\\
        BiM&	9.2&	9.2&	9.2&	\textbf{9.3}&	9.1	&9.4&	\textbf{0.2}\\
        PGD&	7.7&	7.6&	7.9&	7.8&	\textbf{8.1}&	7.7&	\textbf{0.4}\\
        APGD&	4.4&	4.1&	4&	\textbf{4.5}&	4.3&	4.4&	\textbf{0.1}\\
       UP&	49.4&	\textbf{49.9}&	48.1&	48.3&	47.6&	46.6&	\textbf{0.5}\\ 
   
\bottomrule
\end{tabular}
% \end{sc}
\end{tiny}
\end{center}
% \vskip -0.1in
\end{table}

\subsubsection{Adversarial Transferability}
\label{cifar10-transferabilty-results}
%---------------------------------------------------------------------------
% Cifar10 - transferabilty results - densenet121
\begin{table}[H]
\caption{Transferability results on CIFAR-10. The adversarial test sets generated from (surrogate) DenseNet121 and fed to the victim models with different architectures. Accuracies shown are for victim models.}
\label{cifar10-transferability-densenet121}
% \scriptsize
\centering
% \vskip 0.15in
\begin{center}
\begin{tiny}
% \begin{sc}
\begin{tabular}{ lccccccr } 
 \hline
 \toprule 
 Pruning \% & \multicolumn{7}{c}{Surrogate Model: DenseNet121} \\
   L2 & APGD&	BiM	&CW&	DF&	FGSM&	PGD	&UP \\\midrule

   MobileNet & & & & & & \\ \midrule
    0\%	&7.1	&8.2	&78.2&	39.2&	14	&9.6&	12.6  \\
    10\%	&4.5&	6.8&	81.8&	42.4&	11.7&	8.9&	11.3\\
    20\%	&3.8	&7&	81.5&	41.4&	13.5&	8.9&	11.2\\
    30\%&	4.9&	7.8	&80.2	&39.3&	12.2&	9.4	&11.2\\
    40\%	&5.5&	9.2&	80&	39	&12.4&	9.6&	11.2\\
    50\%	&6&	9.7&	77.8&	37&	12.1&	11.2&	11.3\\\midrule
    
    ResNet50 & & & & & & \\ \midrule						
    0\%	&8.6&	9.4&	76.2&	36.9&	18.3&	10.6&	14.2\\
    10\%&	3.8&	6.6&	78.8&	40.2&	17.8&	8.5&	13.2\\
    20\%&	3.7	&9.2&	77.9&	37.9&	18.7&	9.2	&11.6\\
    30\%&	4.1	&5.8&	80.8&	40.4&	17.2&	7.8	&13.5\\
    40\%&	5.8&	8.6	&73.6&	38&	19.1&	12.5&	9\\
    50\%&	7.9	&7.4&	68.7&	34.4&	17&	11.7&	10.6\\ \midrule
    
    VGG19	& & & & & & \\ \midrule					
    0\%	&4.5&	8.8	&78.2&	38.9&	18.7&	9.1&	10.9\\
    10\%	&4.1&	8.4	&80.5&	41.6&	18.2&	9.3	&10.7\\
    20\%&	4.1&	8.2&	80.2&	40.5&	17.9&	10.4&	11.2\\
    30\%&	4.4	&8.5&	80.5&	40.7&	15.8&	10.5&	12.2\\
    40\%&	5.1	&8.8&	79.5&	39.8&	17.5	&10.8	&12.5\\
    50\%&	4.2	&7.5&	81.2&	41.2&	16.7&	3.8	&8.4\\

   \bottomrule
\end{tabular}
\end{tiny}
\end{center}
\end{table}
%---------------------------------------------------------------------------

%---------------------------------------------------------------------------
% Cifar10 - transferabilty results - mobilenet
\begin{table}[H]
\caption{Transferability results on CIFAR-10. The adversarial test sets generated from (surrogate) MobileNet and fed to the victim models with different architectures. Accuracies shown are for victim models.}
\label{cifar10-transferability-mobilenet}
% \scriptsize
\centering
% \vskip 0.15in
\begin{center}
\begin{tiny}
% \begin{sc}
\begin{tabular}{ lccccccr } 
 \hline
 \toprule 
 Pruning \% & \multicolumn{7}{c}{Surrogate Model: MobileNet} \\
   L2 & APGD&	BiM	&CW&	DF&	FGSM&	PGD	&UP \\\midrule

   DenseNet121 & & & & & & \\ \midrule

    0\% &13.3&	8.6	&82.3&	46.5&	16.1&	11.5&	65.2\\
    10\% &  9.8	&9.4&	85.5&	45	&14.5&	8.1&	70.3\\
    20\% & 10.5&	8.5	&84.7&	44.1&	14.5&	8.9	&70.2\\
    30\% & 10.4&	9.8&	85.6&	45.1&	13.2&	8.5&	69.5\\
    40\% &10.9&	6.1	&84.2&	44.6&	13.4&	8.8&	68.7\\
    50\%  & 9.7&	8.4&	79.7&	40.2&	13.8&	8.1&	67.9\\  \midrule
    
    ResNet50 & & & & & & \\ \midrule						
   0\%  &18.4&	9.5&	77&	41.1&	19.7&	15.1&	62.5\\
    10\%& 16.9&	8&	80.3&	45.8&	19.4&	15	&67.3\\
   20\% &15&	7.9&	77.7&	45.5&	19.7&	13.6&	65.3\\
   30\% & 14.8&	6.8&	81.4&	45.2&	20.1	&13.3&	66.5\\
   40\% & 16&	5.8	&74.2&	41	&21.5&	14.7&	61.7\\
   50\% & 15.5	&9.5&	69.6&	39.8&	19.5	&15.7	&56.8\\\midrule
    
    VGG19	& & & & & & \\ \midrule					
    0\% & 16.3&	5.4	&77.7&	42.2&	18.6&	12.9&	67.7\\
   10\%&  14.6&	5.9	&81.1&	44.6&	16.6&	9.2&	68.7\\
   20\%&  17.2&	5.5	&80.4&	44.7&	16.5&	11&	 69.3\\
   30\%&  15.6&	5.6	&80.4&	43.3&	16.8&	10.2&	69.3\\
   40\%&  14.2	&6	&80.3&	43.6&	16.6&	8.8	& 67.1\\
   50\% & 13.7	&6	&80.8&	43.9&	15.1&	10.3&	68\\

   \bottomrule
\end{tabular}
\end{tiny}
\end{center}
\end{table}
%---------------------------------------------------------------------------

%---------------------------------------------------------------------------
% Cifar10 - transferabilty results - Vgg19
% \vspace{-0.5in}
\begin{table}[H]
\caption{Transferability results on CIFAR-10. The adversarial test sets generated from (surrogate) VGG19 and fed to the victim models with different architectures. Accuracies shown are for victim models.}
\label{cifar10-transferability-vgg19}
% \scriptsize
\centering
% \vskip 0.15in
\begin{center}
\begin{tiny}
% \begin{sc}
\begin{tabular}{ lccccccr } 
 \hline
 \toprule 
 Pruning \% & \multicolumn{7}{c}{Surrogate Model: VGG19} \\
   L2 & APGD&	BiM	&CW&	DF&	FGSM&	PGD	&UP \\\midrule

     MobileNet	& & & & & & \\ \midrule	
     0\%	&12.9	&11.3&	74.5&	49.8&	15.9&	15.3&	49.2 \\
    10\%	&11.5&	10.9&	82.1&	53.5&	17.4	&10.6	&53.4\\
    20\%&	10.5&	12.4&	81.6&	51.4&	16.8&	9.7&	52.5\\
    30\%&	11.3&	12.1&	80.9&	50.4&	17.7&	12	&51.9\\
    40\%&	11.5&	12.2&	78.6&	49.4&	16.5&	10.5&	48.5\\
    50\%&	12.5&	13.1&	77.5&	49.2&	15&	11.7&	47.9\\ \midrule
   DenseNet121 & & & & & & \\ \midrule
    0\%&	14.7&	13.1&	81.1&	53.6&	16.5&	15.4&	52.8\\
    10\%&	9.5&	9.9&	85.3&	54.3&	19.4&	10.7&	60.6\\
    20\%&	9.9	&10.2&	84.3&	54.3&	19.2&	11.4	&61.7\\
    30\%&	10.3&	12.2&	84.7&	51.4&	17.4&	11.5&	60\\
    40\%&	12	&13.2	&83.8&	52.4&	16.7&	12.2&	61\\
    50\%&	11.4&	12	&79.1	&48&	16&	12.9&	56.4 \\\midrule
    
    ResNet50 & & & & & & \\ \midrule						
    0\%&	17.1&	12&	77.1&	49.2&	20.8&	17.2&	53.8\\
    10\%&	14.6&	10.7&	79.3&	51.8&	21.5&	11.5&	59.7\\
    20\%&	13.7&	12.3&	77.8&	49.3&	21.5&	12.8&	59.7\\
    30\%&	10	&82.2&	82.2&	56.2&	24.2&	12.5&	59.9\\
    40\%&	14.4&	10.5&	74.1&	43.6&	24.1&	11.1&	54.2\\
    50\%&	16.5	&9.4&	68.9&	41.9&	23.7&	16	&50.7\\

   \bottomrule
\end{tabular}
\end{tiny}
\end{center}
\end{table}

\begin{table}[H]
\caption{Inference time (ms) of base and L2 pruned models on CIFAR-10. Inference time calculated averaged over 100 runs with batch size = 32}
\label{cifar10-inference-time-32}
% \scriptsize
\centering
% \vskip 0.15in
\begin{center}
\begin{tiny}
% \begin{sc}
\begin{tabular}{ lccccr } 
 \hline
 \toprule
 Model &   \multicolumn{4}{c}{Inference Time(ms)} \\
      & Base & 10\% & 30\%  & 50\%   \\ 
 \midrule  
VGG19 & 102.54 $\pm$ 8.35& 	91.30 $\pm$ 4.96 	&94.23 $\pm$ 5.25 &	93.21 $\pm$ 5.49   \\
RN50 &	93.92 $\pm$ 5.50&	93.82 $\pm$ 4.48& 	90.42 $\pm$ 5.17	& 	87.93 $\pm$ 3.63\\
MN  & 20.43 $\pm$ 1.61	&19.41 $\pm$ 2.09&	 	19.65 $\pm$ 1.98&	 	19.60 $\pm$ 1.80 \\
DN121&	63.77 $\pm$ 5.67&	59.39 $\pm$ 3.64	&	59.23 $\pm$ 3.88	& 57.26 $\pm$ 4.10 \\
\bottomrule
\end{tabular}
% \end{sc}
\end{tiny}
\end{center}
% \vskip -0.1in
\end{table}
%---------------------------------------------------------------------------

\subsection{CIFAR-100 Results}
\label{cifar100-results}
Since we are NNCF-based pruning the models for which we initially remove the 10\% unimportant filters and then for a target pruning of 20-50\% it iteratively fine tunes and removes the filters and weights which are unimportant to reach target pruning, and for 10\% of pruning target it just fine-tunes the model, so that leads to the DF attack as shown in table \ref{cifar100-attacks-l2-without-imagenet}, \ref{cifar100-attacks-l1-without-imagenet}, 
 \ref{cifar100-attacks-gm-without-imagenet}, to show significant accuracy change.
% \subsubsection{Pruned models finetuned using base model weights}
% \label{cifar100-results-without-imagenet}

%---------------------------------------------------------------------------
% Cifar100 - various pruning type table - L1, L2, GM
\begin{table}[H]
\caption{Base \& pruned model accuracy on benign CIFAR-100 test set. The pruning target ranges from 10\% - 50\% with $L_1, L_2,$ and Geometric Median (GM) pruning criterion.}
\label{cifar100-pruning-type-without-imagenet}
\centering
% \vskip 0.15in
\begin{center}
\begin{tiny}
% \begin{sc}
\begin{tabular}{ lcccr } 
 \hline
 \toprule
   Pruning  & Pruning 	& \multicolumn{3}{c}{ Benign Test Accuracy } \\
    Type & \% & MN	& DN121  &	RN50 \\ \midrule
    Base&	0\%	&50.33&	49.97&	43.3 \\ \midrule
				
    L1	&10\%&	\color{darkgreen}{\textbf{57.35}}&	62.11&	\color{red}{\textbf{34.22}} \\
    L1	&20\%&	55.79&	60.48&	46.34\\
    L1	&30\%&	54.58&	58.09&	50.68\\
    L1	&40\%&	53.05&	57.04&	50.54\\
    L1	&50\%&	\color{red}{\textbf{51.04}}&	\color{red}{\textbf{53.63}}&	\color{darkgreen}{\textbf{51.3}}\\\midrule

    L2	&10\%&	57.24&	62.07&	39.32\\
    L2	&20\%&	55.41&	61.09&	43.49\\
    L2	&30\%&	54.42&	59.66&	39.83\\
    L2	&40\%&	53.01	&58.27	&44.03\\
    L2	&50\%&	51.4&	53.94	&45.97\\\midrule

    GM&	10\%&	56.95&	\color{darkgreen}{\textbf{62.17}}&	41.59\\
    GM&	20\%&	55.22&	60.9&	44.56\\
    GM&	30\%&	54.51&	59.25&	46\\
    GM&	40\%&	52.97	&58.05	&47.38\\
    GM&	50\%&	51.44&	54.57&	42.6\\\midrule
    Max Delta&	&	{\textbf{7.02}}&	{\textbf{12.2}}&	{\textbf{8}}\\
    Min Delta&	&	{\textbf{0.71}}&	{\textbf{3.66}}&	{\textbf{-9.08}}\\\midrule

\bottomrule 
\end{tabular}
\end{tiny}
\end{center}
\end{table}

%---------------------------------------------------------------------------
% Cifar100 - various attacks table with L2 norm
\begin{table}[H]
\caption{ Base and Pruned model  on CIFAR-100. Adversarial Test Accuracy and \textbf{ L2 filter pruned model} with 10- 50\%  pruning and their Adversarial Test Accuracy are shown. Examples are generated from the base model and fed to the base and pruned models. }
\label{cifar100-attacks-l2-without-imagenet}
\begin{center}
    \begin{tiny}

\begin{tabular}{ lccccccr } 
 \hline
 \toprule
  Attack  & Base  & \multicolumn{5}{c}{L2 Pruned }  & max \\
  &   & 10\%	& 20\%  &30\% & 40\% & 50\% & \(\delta\)\\ \midrule
     
    MobileNet & & & & & & \\ \midrule
    DF&	20.6&	\textbf{35.07}&	34.13&	33.92&	33.25	&32.99&	\textbf{14.47} \\
    FGSM	&7.9&	2.29	&2.45&	2.19&	2.6&	2.49&	-5.3\\
    BiM	&1.5&	0.86&	0.84&	0.77&	0.64&	0.66&	-0.64\\
    PGD&	7.7&	0.95&	0.95&	0.94&	1.21&	1.29&	-6.41\\
    APGD&	8.6&	0.79&	0.86&	0.77&	1.19&	1.07&	-7.41\\
    UP&	25.5&	\textbf{39.62}&	38.96&	36.66&	35.18&	33.67&	\textbf{14.12}\\ \midrule

    DenseNet121 & & & & & & \\ \midrule
   DF&	13.49&	\textbf{35.76}&	35.1&	34.18&	33.64&	31.27&	\textbf{22.27}\\
    FGSM&	1.45&	1.61&	\textbf{1.98}&	1.54&	1.81&	1.2&	\textbf{0.53}\\
    BiM	&0.92&	0.9&	1.01&	\textbf{1.23}&	0.82&	0.82&	\textbf{0.31}\\
    PGD&	3.22&	5.95&	\textbf{6.1}&	5.16&	5.85&	4.07&	\textbf{2.88}\\
    APGD&	1.03&	1&	1&	1.05&	0.96&	\textbf{1.11}&	\textbf{0.08}\\
    UP&	3.75&	6.59&	6.5&	5.6&	\textbf{6.94}&	4.21&	\textbf{3.19}\\\midrule

    ResNet50 & & & & & & \\ \midrule
    DF	&22.1&	29.55&	31.59&	29.2&	31.24	&\textbf{32.94}&	\textbf{10.84}\\
    FGSM&	13.2&	4.85&	2.43&	2.94&	1.81&	2.52&	-8.35\\
    BiM	&1.7	&0.78&	0.79&	0.75&	0.9	&0.76&	-0.8\\
    PGD	&14.7&	\textbf{14.72}&	7.99&	8.41&	6.08&	7	&\textbf{0.02}\\
    APGD&	11.7&	3.56&	1.31&	1.42	&1.21&	1.31&	-8.14\\
    UP	&26.5	&20.93&	16.65&	13.98	&11.84&	13.03&	-5.57\\
    
\bottomrule
\end{tabular}
    \end{tiny}
\end{center}
\end{table}
%---------------------------------------------------------------------------

%---------------------------------------------------------------------------
% Cifar100 - various attacks table with L1 norm
\begin{table}[H]
\caption{Base and Pruned model on CIFAR-100. Adversarial Test Accuracy and \textbf{ L1 filter pruned model} with 10- 50\%  pruning and their Adversarial Test Accuracy are shown. Examples are generated from the base model and fed to the base and pruned models.  }
\label{cifar100-attacks-l1-without-imagenet}
\begin{center}
    \begin{tiny}

\begin{tabular}{ lccccccr } 
 \hline
 \toprule
  Attack  & Base  & \multicolumn{5}{c}{L1 Pruned }  & max \\
  &   & 10\%	& 20\%  &30\% & 40\% & 50\% & \(\delta\)\\ \midrule
     
    MobileNet & & & & & & \\ \midrule
    DF&	20.6&	\textbf{35.2}&	34.14&	34.07&	32.89&	32.37&	\textbf{14.6} \\
    FGSM&	7.9&	2.11&	2.3&	2.37&	2.05&	2.23	&-5.53 \\
    BiM	&1.5&	0.86&	0.81&	0.74&	0.58&	0.7&	-0.64 \\
    PGD	&7.7&	0.74&	0.76&	1.07&	0.99&	1.22&	-6.48 \\
    APGD&	8.6	&0.83&	0.81&	0.99&	0.99&	1.07&	-7.53 \\
    UP	&25.5&	\textbf{39.46}&	37.55&	37	&35.91&	34.64	&\textbf{13.96} \\ \midrule

    DenseNet121 & & & & & & \\ \midrule
    DF&	13.49&	\textbf{35.97}&	34.75&	33.4&	33.3&	30.6&	\textbf{22.48} \\
    FGSM&	1.45&	1.56&	\textbf{1.87}&	1.82&	1.69&	1.11&	\textbf{0.42} \\
    BiM&	0.92&	\textbf{1.25}&	0.97&	0.99&	1.03&	1.06&	\textbf{0.33} \\
    PGD	&3.22&	5.62&	\textbf{5.91}&	5.2&	4.98&	3.75&	\textbf{2.69 }\\
    APGD&	1.03	&1.07&	\textbf{1.13}&	1.2&	1&	0.95&	\textbf{0.17} \\
    UP&	3.75&	5.67&	5.26&	5.61&	\textbf{5.83}&	3.04&	\textbf{2.08} \\\midrule
    
    ResNet50 & & & & & & \\ \midrule
    DF&	22.1&	25.66&	33.56&	36.8&	36.77&	\textbf{37.19}&	\textbf{15.09} \\
    FGSM&	13.2&	4.51&	3.24&	1.64&	2.46&	2.52&	-8.69 \\
    BiM	&1.7&	0.63	&0.8&	0.82&	0.77&	0.71&	-0.88 \\
    PGD	&14.7&	14.17	&8.75&	7.23&	7.62&	7.39&	-0.53 \\
    APGD&	11.7&	3.71&	1.21	&1.26	&1.24&	1.09&	-7.99 \\
    UP&	26.5&	19.9&	15.66&	13.16&	14.23&	15.02&	-6.6\\
    
\bottomrule
\end{tabular}
    \end{tiny}
\end{center}
\end{table}
%---------------------------------------------------------------------------

%---------------------------------------------------------------------------
% Cifar100 - various attacks table with gm norm
\begin{table}[H]
\caption{Base and Pruned model on CIFAR-100. Adversarial Test Accuracy and \textbf{ GM filter pruned model} with 10- 50\%  pruning and their Adversarial Test Accuracy are shown. Examples are generated from the base model and fed to the base and pruned models.  }
\label{cifar100-attacks-gm-without-imagenet}
\begin{center}
    \begin{tiny}

\begin{tabular}{ lccccccr } 
 \hline
 \toprule
  Attack  & Base  & \multicolumn{5}{c}{GM Pruned }  & max \\
  &   & 10\%	& 20\%  &30\% & 40\% & 50\% & \(\delta\)\\ \midrule
     
    MobileNet & & & & & & \\ \midrule
    DF&	20.6&	\textbf{34.83}&	33.89&	33.93&	33.74&	33.45&	\textbf{14.23} \\
    FGSM&	7.9&	2.27&	1.94&	2.43	&2.2&	2.39&	-5.47 \\
    BiM	&1.5&	0.83&	0.84&	0.72&	0.67&	0.68&	-0.66 \\
    PGD	&7.7&	0.87&	0.88&	0.99&	1.12	&1.2	&-6.5 \\
    APGD&	8.6&	0.89&	0.81&	1.09&	1.29&	1.1&	-7.31 \\
    UP&	25.5&	\textbf{39.72}&	37.72&	36.62&	34.61&	34.15&	\textbf{14.22} \\ \midrule

    DenseNet121 & & & & & & \\ \midrule
    DF&	13.49&	\textbf{35.89}&	34.59&	34.01&	33.61&	32.05&	\textbf{22.4} \\
    FGSM&	1.45&	1.65&	\textbf{1.8}&	1.61&	1.4	&1.51	&\textbf{0.35} \\
    BiM	&0.92&	\textbf{1.27}&	0.94&	1.26&	1.24&	0.82&	\textbf{0.35} \\
    PGD	&3.22&	\textbf{6.03}&	5.66&	4.69&	5.62&	5.33&	\textbf{2.81} \\
    APGD&	1.03&	1.06	&0.97&	0.88&	1.16&	\textbf{1.31}&	\textbf{0.28} \\
    UP&	3.75&	6.56&	6.11&	4.65&	\textbf{6.81}&	6.27&	\textbf{3.06}  \\\midrule

    ResNet50 & & & & & & \\ \midrule
    DF&	22.1&	30.69&	31.42&	34.09&	\textbf{34.5}&	30.29&	\textbf{12.4 }\\
    FGSM&	13.2&	3.3&	2.04&	3.95&	3.29	&2.62&	-9.25 \\
    BiM	&1.7	&1.03&	0.88&	0.91&	0.83&	0.75&	-0.67 \\
    PGD	&14.7&	10.18&	6.3&	12.05&	9.15	&7.6&	-2.65 \\
    APGD&	11.7&	1.93&	1.08&	1.88	&1.17&	1.28&	-9.77 \\
    UP	&26.5	&21.67&	11.35&	25.03&	17.9&	15.67&	-1.47\\
    
\bottomrule
\end{tabular}
    \end{tiny}
\end{center}
\end{table}
\begin{table}[H]
\caption{CIFAR-100 -Base and Pruned Model sizes.}
\label{cifar100-size}
\centering
% \vskip 0.15in
\begin{center}
\begin{tiny}
\begin{sc}
\begin{tabular}{ lccr } 
 \hline
 \toprule
 Model &  \multicolumn{2}{c}{Size (MB)}  &  \\
     &  Base  & Pruned &  \%Reduction  \\ 
 \midrule  
ResNet50	&273	&92&	66.30 \\
MobileNet	&39	&13	&66.67 \\ 
DenseNet121	&84&	29	& 65.47 \\ 
\bottomrule
\end{tabular}
\end{sc}
\end{tiny}
\end{center}
% \vskip -0.1in
\end{table}

\end{document}